\documentclass[a4paper,10pt]{article}
\usepackage[english]{babel}
\usepackage[utf8]{inputenc}
\usepackage{comment}
\usepackage{color}
\usepackage{graphicx}
\usepackage{multicol}
\usepackage{blindtext}
\usepackage{hyperref}
\usepackage[export]{adjustbox}

\title{Unsupervised Image to Image Translation}
\author{
    Luca Barras\\ email \href{mailto:luca.barras@epfl.ch}{luca.barras@epfl.ch}
    \and
    Samuel Chassot\\ email \href{mailto:samuel.chassot@epfl.ch}{samuel.chassot@epfl.ch}
    \and
    Daniel Filipe Nunes Silva\\ email \href{mailto:daniel.nunessilva@epfl.ch}{daniel.nunessilva@epfl.ch}
    \and
    Supervised by Deblina Bhattacharjee (team 3)
}
\date{Spring 2021}

\begin{document}

\maketitle
\begin{abstract}
   Unsupervised image-to-image translation methods have received a lot of attention in the last few years. Multiple techniques emerged tackling the initial challenge from different perspectives. Some focus on learning as much as possible from several target style images for translations while other make use of object detection in order to produce more realistic results on content-rich scenes. In this work, we assess how a method that has initially been developed for single object translation performs on more diverse and content-rich images. Our work is based on the FUNIT\cite{FUNIT} framework and we train it with a more diverse dataset. This helps understanding how such method behaves beyond their initial frame of application. We present a way to extend a dataset based on object detection. Moreover, we propose a way to adapt the FUNIT framework in order to leverage the power of object detection that one can see in other methods.
\end{abstract}

\section{Introduction}

We explore unsupervised image-to-image (I2I) translation techniques that allow us to transform images from a given domain by applying characteristics of images from an other domain (e.g., turn an rainy landscape picture into a sunny one).

Although state-of-the-art frameworks perform reasonably well when working with realistic styles, they produce unusable results when applied on unrealistic ones (e.g., cartoons or comics). They also struggle to apply style to images composed of complex scenes with multiple objects, especially when the style image's class was not seen at training time.

The goal of this project is to explore how I2I frameworks behave when facing unforeseen target classes.

We base our work on three main papers: FUNIT, INIT\cite{INIT} and DUNIT\cite{DUNIT}. We use the framework of FUNIT with the dataset from INIT. This framework takes advantage of having multiple target domain style images at its disposable to extract the most meaningful characteristics to be applied on the input. However, we do not know how this framework may perform when translating content-rich scenes.

The first step is to assess FUNIT's performance on INIT's dataset, which is known for being diverse and challenging. Then, we explore how we could improve FUNIT capabilities on content-rich images by extending the INIT dataset. Finally, we suggest how we could modify the FUNIT using techniques used by INIT and DUNIT that leverage object detection.

The main contributions of our work are the following:
\begin{itemize}
    \item We assess FUNIT performance with INIT dataset.
    \item We generate a new dataset base on INIT's one by running an object detector on it.
    \item We present a way to incorporate an object detector in FUNIT.
\end{itemize}

\section{Related Work}

\subsection{Few-Shot Unsupervised Image-to-Image Translation (FUNIT)}
The particularity of this framework is that it uses multiple domains from the training set at the training time. Then it only needs a few images of a possibly unseen domain during the test time to generate the output image. The goal of FUNIT is to map an image of a source class to an analogous image of an unseen target class by leveraging a few target images that are made available at the test time.

To achieve this goal, it uses a particular generator composed of an encoder for the content image (from which it extracts structural information) and a second encoder for the style images (from which it extracts style information). Then, a decoder takes the 2 encoded representation as its input. The encoder for the style images can take 1 to N image instances of the same class as input and merges their latent vector representations by taking the average. The decoder then, merges the style code to the content code, by taking the affine transformation of the features from a series of fully connected layers. The affine transformation acts globally on an image, thereby preserving its structure.

Although the framework handles instance translation, given a limited number of such instances, it has two main limitations. First, a very disparate domain from the ones seen during training significantly affects the results obtained. Second, the model performs well, mainly on images where the object is well identified. Thus, a content rich scene with multiple objects, doesn't translate well under this model.

\subsection{Towards Instance-level Image-to-Image Translation}
This paper proposes a solution to tackle one problem of the existing I2I translation tools: they give bad results when the content image is a complex scene with multiple distinct objects. The authors propose to use different styles representations for the background, objects and global image. To do that, they add a network to recognize objects in the scene. Then, they use two different encoders/decoders pairs instead of one: one for the global image and one for the objects. Each encoder produces a pair of feature vectors: one for the content and one for the style (e.g., for the global image: $E_g(I) = (c_g, s_g)$). Each decoder takes two feature vectors (for content and style) and produces an image. The images produced by the two decoders are then merged to form the final result.

The major limitation of this model is the need of an annotated dataset with the location of the bounding boxes around objects of the scene. Indeed, it takes a lot of time to annotated thousands of images. In our project, we explore a way of automatically detect objects to overcome this limitation.

\subsection{Detection-based Unsupervised Image-to-Image Translation (DUNIT)}
DUNIT is fundamentally different from FUNIT: DUNIT uses only one style images to learn the style's characteristics where FUNIT extracts them from multiple target style images. DUNIT, just as INIT, treats objects and background separately, where FUNIT treats the image as a whole. DUNIT brings some improvements that allows it to outperform state-of-the-art methods according to multiple metrics, giving the closest similarity for the translations of the three models, quantitatively and qualitatively. 
 
 DUNIT keeps track of instance-level information and uses it for testing. It is different from INIT which treats the generated images as a whole. DUNIT's architecture is built on the DRIT\cite{DRIT}'s backbone, which separates to content from the style, therefore learning style-invariant representations. DUNIT also introduces a new loss function in order to enforce consistency between the input and the output images. This basically makes sure that the detected objects are present in both image at the same coordinates. This idea could be extended as suggested in the DUNIT's paper conclusion to enforce not only objects position but also their pose. The following references may provide useful material to explore this direction \cite{li2020deformation} and \cite{papaioannidis2020domain}.

\section{Methodology}
\subsection{Model and Architecture}
We now present the FUNIT framework that we use.

\begin{figure}
  \includegraphics[width=\linewidth]{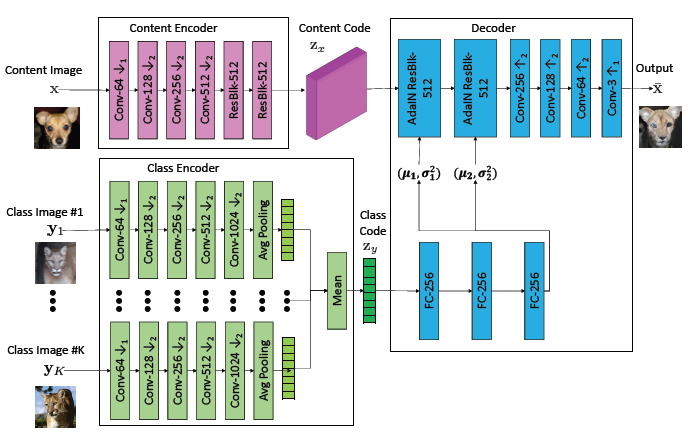}
  \caption{generator architecture (from FUNIT)}
  \label{fig:generator_architecture}
\end{figure}

\textbf{Generator:} The generator takes as input one content image and a set of N style images that belong to the same class. This generator is composed of two encoders, one for the content and one for the style, and one decoder. The generator architecture is illustrated in Fig.\ref{fig:generator_architecture}. The content encoder maps its input image to a content latent code which represents information about the image structure. The style encoder maps its inputs to intermediate style latent code that represents the style of each style image. The global style latent code is then the natural mean of all these style latent codes. The content encoder is composed of several 2D convolution layers followed by residual blocks. The style encoder is composed of several 2D convolution layers. The decoder takes two inputs, the content latent code and the style latent code. The content latent code goes through some adaptive instance normalization residual blocks followed by several upscale convolution layers. The style latent code goes through several fully connected layers to regress in AdaIN parameters. These parameters allow to modify globally the features maps in the AdaIN residual blocks. To change the style of an image, we want to apply a global transformation on it to preserve its general structure.

\textbf{Discriminator:} The discriminator has as many outputs as there are classes and each of these outputs is a binary classification that indicates whether the input is real or fake with respect to the class. During the training, only of the output given by the class of the input image is used (for each translation). For example if the input style image is of the class "rainy", the model only uses the output corresponding of the class rainy in the discriminator and updates the parameters with respect to this output. It penalizes the discriminator when it is wrong, i.e., the input image is fake and the discriminator outputs true (aka real) for the image's class or the other way around.

\subsection{Training}
We use the same loss functions as FUNIT.
There are three different loss functions in the model: the GAN loss, the content reconstruction loss and the feature matching loss. To train the model we solve the following minimax optimization problem:

\begin{equation}
\displaystyle \min_{D}\max_{G} \mathcal{L}_{GAN}(D, G) + \lambda_{R}\mathcal{L}_{R}(G) + \lambda_{F}\mathcal{L}_{FM}(G)
\end{equation}

Here, $\mathcal{L}_{GAN}$ is the GAN loss, $\mathcal{L}_R$ is the content reconstruction loss and $\mathcal{L}_{FM}$ is the feature matching loss.

The GAN loss is given by 
\begin{equation}
\mathcal{L}_{GAN}(G, D) = E_x[-logD^{c_x}(x)] + E_{x, \{y_1, \dots,y_k\}}[log(1 - D^{c_y}(\bar{x}))]
\end{equation}

This loss penalizes when the discriminator gets a real data of source class $c_x$ and its $c_x$th output is false or when it gets a fake data of source class $c_x$ and its $c_x$th output is true. So this loss allow the discriminator to improve the discrimination between real and fake data, and the generator to improve fooling the discriminator in generating images closer to real images.

The content self reconstruction loss is given by
\begin{equation}
\mathcal{L}_R(G) = E_x[||x - G(x, \{x\})||_1^1]
\end{equation}
Minimizing this loss encourages the generator to generate an output identical to the input when the input content image and input class image are the same.

And finally we have the feature matching loss given by

\begin{equation}
\displaystyle \mathcal{L}_F(G) = E_{x,\{y_1,\dots,y_k\}}[||D_f(\bar{x})-\sum_{k} \frac{D_f(y_k)} {K}||_1^1]
\end{equation}

This loss allows to regularize the training. In this formula, $D_f$ is the discriminator model without the last prediction layer and it is used to extract features. $\bar{x}$ is the output generated by the discriminator. $D_f$ allows to extract features from the translation output $\bar{x}$ and the different class images $\{y_1,\dots,y_k\}$.

\section{Experiments}
\subsection{Dataset}
For the following experiments, we use the dataset of INIT. The dataset contains street images from Tokyo, Japan, captured by a SEKONIX AR0231 camera mounted on a car. The images have been captured from different time of day and different weather which allows to create 4 different classes: sunny, cloudy, night and rainy. The original images are in 3000x4000 pixels but here we will use only the 1280x1920 pixels versions considering computing power limitations. Furthermore, we reduce the size of the dataset by removing some images for the same reason. We assume that running multiple trainings with less images provides more convincing results than running fewer trainings with the full dataset. Initially there are 155,529 images across the 4 classes, cloudy, sunny, night and rainy. We keep only $\frac{1}{3}$ of the original dataset considering storage limitations. As cloudy and sunny are the most populated classes, we mainly remove images from them to balance the available data for each class. We end up with the number of images for each class shown in table~\ref{tab:images_per_class}.

\begin{table}[]
    \centering
    \begin{tabular}{|c|c|}
    \hline
        class & \#images\\
        \hline
        cloudy & 12723\\
        sunny & 10678\\
        rainy & 2226\\
        night & 6705\\
    \hline
    \end{tabular}
    \caption{Number of images per class in our INIT dataset subset}
    \label{tab:images_per_class}
\end{table}

\subsubsection{RetinaNet}
At first, we use the dataset with only the 4 classes of the INIT dataset. As emphasized by the authors of FUNIT, the more different classes are available at training the better are the results. So we use RetinaNet\cite{Retinanet}, a single shot object detector, to extract objects from the images and create new classes. Moreover, having these objects as individual images can increase the quality of the results for complex scenes. When running RetinaNet on the dataset, we use a threshold of 0.5 to keep only the results with more than 50\% of confidence. For each type of object $Obj$ extracted from an image of the class $C$, we create a class $C\ Obj$. For example, if we extract a car from an image of the sunny domain, we will get the new "sunny - car" class. After this operation, we get 178 different classes. Some of them look however like noise (e.g., boomerang or giraffe), and are certainly due to some imprecision of the object detector. So after manual checks of the images in these classes, we conclude that the images are not representing the objects of those classes. We then decide to remove them and keep only 97 classes (23 in rainy, 24 in night, 25 in cloudy, 25 in sunny). See Appendix for detailed list of the produced classes.

\subsection{Implementation}
For the implementation, we use RMSProp with a learning rate of $0.0001$ as optimizer. We set the hyper-parameters $\lambda_R = 0.1$ and $\lambda_F = 1$. The batch size we used vary according to the experiments. We have to use small batch size between 1 and 8 because of the limited computing power of our machine.

\subsection{Experimentation}
We proceed in multiple phases with different configurations. The really big size of the dataset represents a challenge to work with on our hardware. Therefore, we have to adapt our dataset and parameters accordingly.

All the following trainings are performed by running the code of FUNIT\footnote{Available on our fork: https://github.com/samuelchassot/FUNIT}.

At each run, we try different batch sizes to use the highest possible given the computational power at our disposal at the time (especially GPU memory). We give the sizes used for the trainings presented in the results.

\textbf{Phase 0:} Due to performance limitations, we only keep a subset of the dataset by picking images from each classes equally. This gives 7440 images in the training set and 1861 images in the test set. Unfortunately, the way we split the dataset for this run is incorrect: we use images from all classes for training and testing which means that the model could already learn each class' distribution. This does not give any exploitable results to measure the performance of FUNIT on unseen domain. However this first run gives encouraging results and proves that the setup is working.

\textbf{Phase 1:} We now split the dataset in a valid way according to the FUNIT's specifications. Classes should not appear in both the test and training set. Therefore, we keep the rainy class for the test and use the others for training. Our biggest concern is that FUNIT's performance increases with the number of distinct classes appearing at training time. However, INIT's only provides 4 classes which is one order of magnitude less than what is presented in the original FUNIT paper. As expected results for this run look terrible when compared to the first one despite more training. This is due to the fact that not all classes are seen in the training set. Nevertheless, this model seems to be learning some low level components of the images. As a whole they do not mean much but when inspected carefully they reveal elements such as lanes, cars or trees shadows. At this stage, it is very delicate to say anything about the performance but it seems encouraging.

\textbf{Phase 2:} As the first phase may hint when compared to the second one, data is key. Therefore, we use the whole dataset we keep (as described in the section about dataset). While experimenting with this dataset, we notice that the classes are not always very accurate, i.e. some images classified as cloudy actually look sunnier that images classified as sunny. Therefore, during this phase, we perform two trainings. The first one keeps rainy as the class in the test set but uses more data and less iterations that during the second run. The other training uses the night class for testing. We expect it to give better results as the night class is the most singular among the 4 classes we have.

\textbf{Phase 3 and RetinaNet:} We try to balance the number of images in each class as it is uneven between classes (e.g., cloudy has a lot more images than night). We so keep all the images for the least populated class (rainy) and keep around the same number of images (2226) in the three other classes (selected randomly). The results are less good than with all the images. This once again indicates than the more is the better concerning data in this setup. During this phase, we also perform a training with the classes produced using RetinaNet. We keep all the night classes in the test set and put all the others in the training set. The training with RetinaNet classes gives promissing results.

\section{Results}
After analyzing manually the produced results, we decided to run quantitative analyses on three of our training:
\begin{itemize}
    \item \textbf{Model 1, "Balanced":} night class as test class, with the balanced number of images in each class.
    \item \textbf{ Model 2, "Unbalanced":} night class as test class, with the whole dataset.
    \item \textbf{Model 3, "RetinaNet":} classes produced with RetinaNet, all the night classes as test classes, with the whole dataset.
\end{itemize}

Model 2 and Model 3 are trained twice to then compute the average and give more representative results. Model 1 gives terrible results and we decide not to run it again and concentrate ourselves on the two other more promising models. See Table~\ref{tab:results_trainings} for details about the runs.

Along with these runs of our own, we fine-tune the FUNIT's pretrained model\footnote{https://github.com/samuelchassot/FUNIT/README}. To do so, we resume a training of the provided model with our dataset but with a learning rate divided by 10 (i.e., 0.00001) for 250,000 iterations. This should keep  most of the learned data from the original training (trained over animal faces) and add a small adaptation layer to make it run on our dataset.

We generate images translated to night (i.e., the only style that the model does not see at training time). We do it in 2 different batches: one inputting 2 style images and one with 5 style images. In order to evaluate the performance of our models, we randomly pick 20 images from each of the training classes, then generate 5 pairs of images using 2 (respectively 5) randomly picked target style images from the night class.

We use the following metrics to assess the performance of our models:
\begin{itemize}
    \item \textbf{Inception Score (IS)} which quantifies the diversity across all the translated outputs from a given class.
    \item \textbf{LPIPS distance (LPIPS)} which quantifies the diversity between translated outputs and seems strongly correlated with the human perception.
\end{itemize}
The results are presented in the tables \ref{tab:measure2} and \ref{tab:measure5}.

We use the fine-tuned FUNIT model as a baseline. All its translations seem to be the same. In other words, our fine-tuned model struggles to generated meaningful new images. We cannot even guess any low-level characteristics like roads or the sky from them. The quantitative results confirm that with a $0.000$ LPIPS and fairly low IS too. This is maybe due to the fundamental difference between the dataset this model had been trained for and ours.

The next observation we make goes against our initial thoughts. In fact, we expected that providing more target style images would perform better as the model could extract more information from the night unforeseen class. However, the overall 2 target style images seems to give slightly better results than 5 target style images but the reality is actually trickier. For our better models, namely model 2 "unbalanced" and model 3 "RetinaNet", 2 target style images give a greater LPIPS but a lower IS compared to 5 target style images. We interpret this as follows: providing more target style images gives qualitatively better results (which explains a greater IS) but the pairs of translated outputs are more averaged and therefore closer to each others (which explains a lower LPIPS).

The observations between the different models seem consistent independently from the number of target style images.

Our model 1 performs barely better than our baseline from a quantitative perspective. However, the translated outputs present some interesting low-level characteristics. Although the overall images are meaningless we can still spot some cars or road markings appearing. This lets us think that our models is actually learning how to extract the content of the scene.

Model 2 "unbalanced" and model 3 "RetinaNet" show much more interesting results. The generated outputs already look closer to real images. We deduct that using more data leads to better results. We also observe that using unbalanced data across the different classes also lead to biased results. In fact, classes whose set are larger eventually translated to more realistic outputs. Also, not all translated images look like dark night since the target style night images may represent different night moments from sunset to sunrise. Finally, it is highly likely that using a greater number of classes helps FUNIT during the training phase despite some gain being due to the amount of data brought by RetinaNet. Also note that RetinaNet qualitatively generates more results that look like actual night. We present some translated outputs in the appendix.

\begin{table}[]
    \centering
    \small\addtolength{\tabcolsep}{-0.1pt}
    \begin{tabular}{|c|c|c|c|}
        \hline
        Model & Test class(es) & Train class(es) & number iterations \\
        \hline
         1: Balanced & Night & Cloudy, sunny, rainy & 483,000\\
         2: Unbalanced & Night & Cloudy, sunny, rainy & 470,000 and 500,000\\
         3: RetinaNet & All night & All cloudy, sunny, rainy & 500,000 and 500,000\\
         \hline
    \end{tabular}
    \caption{Detailed information about the runs}
    \label{tab:results_trainings}
\end{table}

\begin{table}[]
\small\addtolength{\tabcolsep}{-4.5pt}
\begin{tabular}{|l|c|c|c|c|c|c|c|c|}
\hline
\begin{tabular}[c]{@{}l@{}}To night:\\ With 2 target\\ images\end{tabular} & \multicolumn{2}{c|}{\textbf{Baseline}} & \multicolumn{2}{c|}{\textbf{Model 1}} & \multicolumn{2}{c|}{\textbf{\begin{tabular}[c]{@{}c@{}}Model 2\ (2 runs)\end{tabular}}} & \multicolumn{2}{c|}{\textbf{\begin{tabular}[c]{@{}c@{}}RetinaNet\ (2 runs)\end{tabular}}} \\ \hline
Metric & \textbf{LPIPS} & \textbf{IS} & \textbf{LPIPS} & \textbf{IS} & \textbf{LPIPS} & \textbf{IS} & \textbf{LPIPS} & \textbf{IS} \\ \hline
From cloudy & 0.000 & 0.444 & 0.001 & 0.445 & 0.093 & 0.709 & 0.314 & 0.764 \\ 
From rainy & 0.000 & 0.463 & 0.001 & 0.484 & 0.114 & 0.594 & 0.316 & 0.767  \\ 
From sunny & 0.000 & 0.449 & 0.001 & 0.452 & 0.076 & 0.693  & 0.320 & 0.872 \\ \hline
\textbf{Average} & 0.000 & 0.452 & 0.001 & 0.460 & 0.095 & 0.666 & \textbf{0.317} & \textbf{0.801} \\ \hline
\end{tabular}
\caption{LPIPS and IS for different models (with 2 targets images)}
\label{tab:measure2}
\end{table}

\begin{table}[]
\small\addtolength{\tabcolsep}{-4.5pt}
\begin{tabular}{|l|c|c|c|c|c|c|c|c|}
\hline
\begin{tabular}[c]{@{}l@{}}To night:\\ With 5 target\\ images\end{tabular} & \multicolumn{2}{c|}{\textbf{Baseline}} & \multicolumn{2}{c|}{\textbf{Model 1}} & \multicolumn{2}{c|}{\textbf{\begin{tabular}[c]{@{}c@{}}Model 2\ (2 runs)\end{tabular}}} & \multicolumn{2}{c|}{\textbf{\begin{tabular}[c]{@{}c@{}}RetinaNet\ (2 runs)\end{tabular}}} \\ \hline
Metric & \textbf{LPIPS} & \textbf{IS} & \textbf{LPIPS} & \textbf{IS} & \textbf{LPIPS} & \textbf{IS} & \textbf{LPIPS} & \textbf{IS} \\ \hline
From cloudy & 0.000 & 0.388 & 0.001 & 0.397 & 0.057 & 0.622 & 0.212 & 0.811 \\ 
From rainy & 0.000 & 0.436 & 0.001 & 0.448 & 0.072 & 0.665 & 0.209 & 0.792 \\ 
From sunny & 0.000 & 0.483 & 0.000 & 0.479 & 0.036 & 0.726 & 0.200 & 0.822 \\ \hline
\textbf{Average} & 0.000 & 0.436 & 0.001 & 0.441 & 0.055 & 0.671 & \textbf{0.207} & \textbf{0.808} \\ \hline
\end{tabular}
\caption{LPIPS and IS for different models (with 5 targets images)}
\label{tab:measure5}
\end{table}

\begin{figure}
  \includegraphics[width=\linewidth]{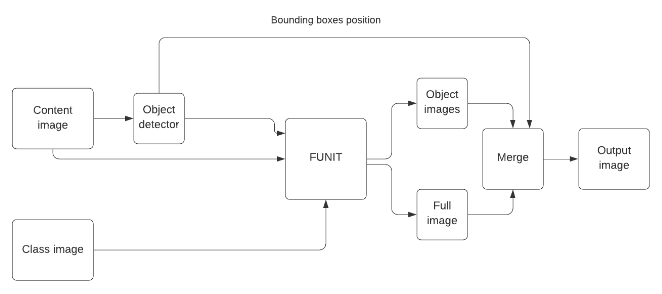}
  \caption{Future architecture with FUNIT}
  \label{fig:architecture_1}
\end{figure}

\begin{figure}
  \includegraphics[width=\linewidth]{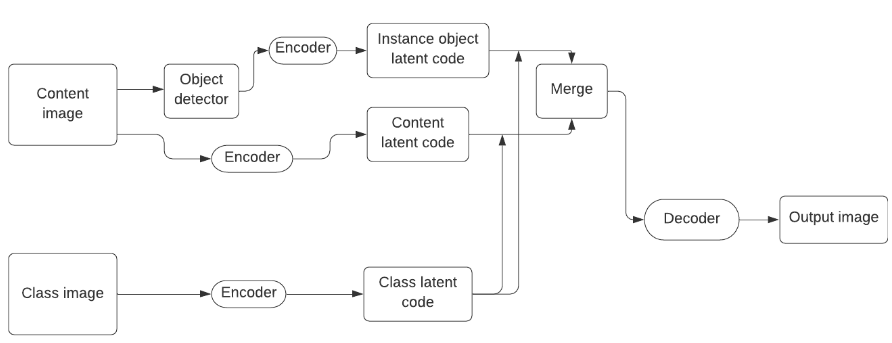}
  \caption{Future architecture in FUNIT}
  \label{fig:architecture_2}
\end{figure}

\subsection{Future work}
We present two new model architectures to increase the results accuracy. We privileged the experimentation with our current setup rather than implementing a new variant of FUNIT. This would have resulted in just few tests of each architecture models which would not have been useful. The first architecture adds an object detector before  the FUNIT model and sends each instance object and the global image in the FUNIT model. It then merges them (replacing at each initial location in the global image the different instance objects) after they passed through the FUNIT model. This architecture is presented on Figure~\ref{fig:architecture_1}. The second model architecture is inspired from the DUNIT model. Instead of merging the extracted objects with the global image after them passing through FUNIT, it merges directly the latent vector inside FUNIT architecture like shown in Figure~\ref{fig:architecture_2}.

\section{Conclusion}
In this project, we train the FUNIT model on a more diverse and complex dataset with fewer different classes. The dataset contains images with complex scenes with multiple objects which is really different from the dataset used in the original FUNIT paper. The results are not promising with the plain dataset but we manage to enhance it and thus get better results by running an object detector first. This creates more classes and therefore helps FUNIT during training. We also observe that the computational power and time are crucial components of such projects (we cannot run with more data or with bigger batches because of the computer we have at our disposal). We however observe that the more data we use, the better get the results. The data and the training time is then a first class citizen for I2I translation. We think that the quality of our results can be enhanced by using a bigger part of the dataset (and extracting objects using RetinaNet) and train the model for a longer duration and with bigger batches.

\newpage

\section{Appendix}
\subsection{Classes produced with RetinaNet}

\begin{multicols}{3}
\noindent sunny - truck \\
sunny - car \\
sunny - traffic light \\
sunny - person \\
sunny - potted plant \\
sunny - motorcycle \\
sunny - stop sign \\
sunny - bench \\
sunny - bus \\
sunny - bicycle \\
sunny - train \\
sunny - vase \\
sunny - tv \\
sunny - dog \\
sunny - clock \\
sunny - chair \\
sunny - parking meter \\
sunny - handbag \\
sunny - toilet \\
sunny - fire hydrant \\
sunny - boat \\
sunny - backpack \\
sunny - suitcase \\
sunny - bird \\
sunny - bottle \\
rainy - backpack \\
rainy - bench \\
rainy - bicycle \\
rainy - bird \\
rainy - boat \\
rainy - bottle \\
rainy - bus \\
rainy - car \\
rainy - chair \\
rainy - clock \\
rainy - fire hydrant \\
rainy - handbag \\
rainy - motorcycle \\
rainy - parking meter \\
rainy - person \\
rainy - potted plant \\
rainy - stop sign \\
rainy - suitcase \\
rainy - toilet \\
rainy - traffic light \\
rainy - train \\
rainy - truck \\
rainy - vase \\
cloudy - car \\
cloudy - truck \\
cloudy - train \\
cloudy - bench \\
cloudy - bus \\
cloudy - traffic light \\
cloudy - person \\
cloudy - motorcycle \\
cloudy - stop sign \\
cloudy - potted plant \\
cloudy - fire hydrant \\
cloudy - suitcase \\
cloudy - bicycle \\
cloudy - backpack \\
cloudy - handbag \\
cloudy - clock \\
cloudy - parking meter \\
cloudy - chair \\
cloudy - toilet \\
cloudy - boat \\
cloudy - dog \\
cloudy - bottle \\
cloudy - tv \\
cloudy - bird \\
cloudy - cell phone \\
night - car \\
night - bus \\
night - truck \\
night - person \\
night - motorcycle \\
night - tv \\
night - traffic light \\
night - train \\
night - bicycle \\
night - fire hydrant \\
night - bench \\
night - clock \\
night - stop sign \\
night - bottle \\
night - parking meter \\
night - potted plant \\
night - vase \\
night - handbag \\
night - dog \\
night - chair \\
night - boat \\
night - laptop \\
night - cell phone \\
night - backpack \\

\end{multicols}

\newpage

\begin{figure}
    \begin{tabular}{cccc}
    To night & from cloudy & from rainy & from sunny\\
    Baseline & \includegraphics[width=.2\linewidth,valign=m]{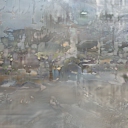} & \includegraphics[width=.2\linewidth,valign=m]{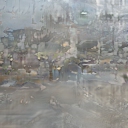} & \includegraphics[width=.2\linewidth,valign=m]{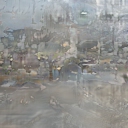}\\
    Model 1 & \includegraphics[width=.2\linewidth,valign=m]{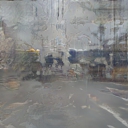} & \includegraphics[width=.2\linewidth,valign=m]{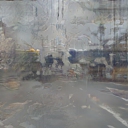} & \includegraphics[width=.2\linewidth,valign=m]{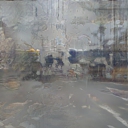}\\
    Model 2 & \includegraphics[width=.2\linewidth,valign=m]{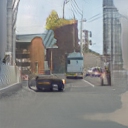} & \includegraphics[width=.2\linewidth,valign=m]{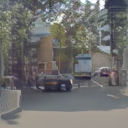} & \includegraphics[width=.2\linewidth,valign=m]{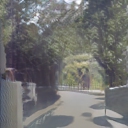}\\
    RetinaNet & \includegraphics[width=.2\linewidth,valign=m]{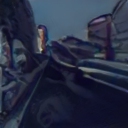} & \includegraphics[width=.2\linewidth,valign=m]{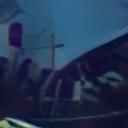} & \includegraphics[width=.2\linewidth,valign=m]{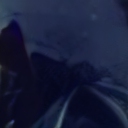}\\
    \end{tabular}
    \caption{Example of translations for every model and every class with 2 target style images}
\end{figure}

\clearpage

\bibliographystyle{unsrt}
\bibliography{references}

\end{document}